\documentclass[times, review, 10pt]{elsarticle}

\usepackage{epsfig}
\usepackage{amssymb}
\usepackage{amsmath}
\usepackage{amsthm}
\usepackage{lineno}
\usepackage{soul}
\usepackage{xcolor}
\usepackage{multirow}
\usepackage{multicol}
\usepackage{booktabs}

\journal{arxiv}

\begin{document}
\begin{frontmatter}

\title{AHMSA-Net: Adaptive Hierarchical Multi-Scale Attention Network for Micro-Expression Recognition}

\author[1]{Lijun Zhang}

\author[1]{Yifan Zhang}

\author[1]{Weicheng Tang}

\author[1]{Xinzhi Sun}

\author[1]{Xiaomeng Wang}

\author[1,2]{Zhanshan Li\corref{cor1}}
\cortext[cor1]{Corresponding author}

\affiliation[1]{organization={College of Computer Science and Technology},
            addressline={Jilin University}, 
            city={Changchun},
            state={Jilin},
            postcode={130012}, 
            country={China}}
            
\affiliation[2]{organization={Key Laboratory of Symbolic Computation and Knowledge Engineering of Ministry of Education},
            addressline={Jilin University}, 
            city={Changchun},
            state={Jilin},
            postcode={130012}, 
            country={China}}

\begin{abstract}
Micro-expression recognition (MER) presents a significant challenge due to the transient and subtle nature of the motion changes involved. In recent years, deep learning methods based on attention mechanisms have made some breakthroughs in MER. However, these methods still suffer from the limitations of insufficient feature capture and poor dynamic adaptation when coping with the instantaneous subtle movement changes of micro-expressions. Therefore, in this paper, we design an Adaptive Hierarchical Multi-Scale Attention Network (AHMSA-Net) for MER. Specifically, we first utilize the onset and apex frames of the micro-expression sequence to extract three-dimensional (3D) optical flow maps, including horizontal optical flow, vertical optical flow, and optical flow strain. Subsequently, the optical flow feature maps are inputted into AHMSA-Net, which consists of two parts: an adaptive hierarchical framework and a multi-scale attention mechanism. Based on the adaptive downsampling hierarchical attention framework, AHMSA-Net captures the subtle changes of micro-expressions from different granularities (fine and coarse) by dynamically adjusting the size of the optical flow feature map at each layer. Based on the multi-scale attention mechanism, AHMSA-Net learns micro-expression action information by fusing features from different scales (channel and spatial). These two modules work together to comprehensively improve the accuracy of MER. Additionally, rigorous experiments demonstrate that the proposed method achieves competitive results on major micro-expression databases, with AHMSA-Net achieving recognition accuracy of up to 78.21\% on composite databases (SMIC, SAMM, CASMEII) and 77.08\% on the CASME\^{}3 database.
\end{abstract}




\end{frontmatter}

\section{Introduction}
\label{sec:intro}
Recognition of micro-expression actions has shown great potential in various real-world applications, including biometric identification \cite{saeed2021facial}, disease detection \cite{huang2021elderly,chen2023catching}, and deception detection \cite{takalkar2020deciphering}. Micro-expressions are involuntary facial movements that often reveal an individual's true emotions \cite{yap2018facial}. However, due to the extremely brief duration of micro-expressions and their occurrence in specific regions of the face, recognizing them with the naked eye is very challenging. Therefore, developing a method that can accurately and efficiently identify micro-expression emotional information is of great significance.

Optical flow is currently the mainstream method for extracting features from micro-expression sequences \cite{liong2019shallow,gan2019off,zhao2019convolutional}. Due to its sensitivity to small, temporal, and non-rigid movements, as well as its independence from specific lighting conditions, optical flow is particularly well-suited for capturing the dynamic changes of micro-expressions \cite{wang2022micro}. However, existing optical flow methods used for micro-expression recognition not only require high-quality input images but also suffer from low sensitivity, often resulting in highly blurred optical flow feature maps \cite{xia2021motion}. To address this issue, we first localized, cropped, and aligned the original micro-expression databases, preemptively removing a large amount of redundant background information. Then, using the onset and apex frames of the micro-expression sequences, we extracted three-dimensional (3D) optical flow feature maps containing horizontal optical flow, vertical optical flow, and optical flow strain, with each dimension learning a specific type of optical flow information. Ultimately, we obtained high-precision, high-quality optical flow feature maps.

In recent years, with the development of deep learning technology, deep learning methods based on attention mechanisms have achieved some breakthrough progress in micro-expression recognition \cite{zhou2023dual}. However, existing attention-based micro-expression recognition methods often rely on fixed feature block sizes, which may result in insufficient capture of details \cite{wang2023shallow}. Additionally, these methods typically utilize only single-scale feature information, neglecting the synergistic effects of multi-scale features, and thus fail to comprehensively represent the complex variations of micro-expressions \cite{wang2024htnet}. These limitations lead to lower accuracy and flexibility when the network is applied to micro-expression recognition tasks. Therefore, in this paper, we design and implement an Adaptive Hierarchical Multi-Scale Attention Micro-Expression Recognition Network (AHMSA-Net) to overcome the limitations of existing methods.

In summary, the main contributions of this paper are as follows:

1. We utilized a high-precision three-dimensional (3D) optical flow method to extract micro-expression features. Unlike traditional optical flow methods that only calculate the horizontal ($u$) and vertical ($v$) components of the optical flow field displacement, we also compute a third component: optical strain ($os$). These three dimensions-$u$, $v$, and $os$-form a high-precision 3D optical flow map.

2. Within the AHMSA-Net framework, we designed a hierarchical attention framework based on adaptive downsampling, which dynamically adjusts the size of the optical flow feature map at each layer. This framework captures subtle micro-expression changes at different granularities, with lower-level attention mechanisms focusing on fine-grained features in local regions and higher-level attention mechanisms targeting coarse-grained features in global regions. This hierarchical attention framework plays a crucial role in recognizing the subtle, localized movements of micro-expressions.

3. We propose a multi-scale attention mechanism that enables AHMSA-Net to learn micro-expression action information by fusing optical flow features from different scales (channel and spatial). The channel attention mechanism assigns varying feature mapping weights to different dimensions of the optical flow feature patches, while the spatial attention mechanism captures the complex interactions between different optical flow patches. This multi-scale attention mechanism effectively highlights key features while suppressing redundant ones.

4. Experimental results demonstrate that the proposed method achieves highly competitive results on major micro-expression databases, with AHMSA-Net achieving a recognition accuracy of up to 78.21\% on composite micro-expression databases (SMIC, SAMM, CASMEII) and 77.08\% on the CASME\^{}3 micro-expression database.

\section{Related Work}

This section summarizes the common representation methods for micro-expression features and some research on the application of attention mechanisms in micro-ex-pression recognition.

\subsection{Common Representation Methods for Micro-Expression Features}
\label{subsec:repre}

The common representation methods for micro-expression features mainly include three categories: texture features, Action Unit (AU) features, and optical flow features. For example, Huang et al. \cite{huang2017discriminative} proposed the Discriminative Spatiotemporal Local Binary Pattern with Rigid Image Preprocessing (DISTLBP-RIP) method for micro-expression recognition by combining shape attributes with dynamic texture features. Secondly, Wang et al. \cite{wang2015micro} defined 16 regions of interest (ROI) based on AU and completed the micro-expression recognition task by extracting features from each ROI. Finally, Chen et al. \cite{chen2022block} proposed the Block-Based Convolutional Neural Network (BDCNN), which uses four optical flow features calculated from the onset and apex frames for micro-expression recognition. Similar methods are adopted in \cite{xia2020revealing,li2021multi}, where optical flow from onset and apex frames is used as network input, reducing the complexity of model input while maintaining good recognition performance.

However, the first method involving texture features has a high dimensionality, often leading to high computational complexity in micro-expression recognition; the second method involving AU features often requires predefined AUs, and different annotators may provide varying annotations, limiting the generalization ability of AU-based recognition methods; in contrast, the third method, based on optical flow features, can reduce feature dimensionality while maintaining good recognition performance, making optical flow the mainstream feature representation method at present.

\subsection{Application of Attention Mechanisms in Micro-Expression Recognition}
\label{subsec:application}

Attention mechanisms play a crucial role in human perception analysis \cite{itti1998model,corbetta2002control}. In recent years, attention mechanisms have been widely applied to various visual tasks \cite{cao2015look,sonderby2015recurrent,zhao2022me} and have achieved state-of-the-art results.

Currently, in micro-expression recognition tasks, researchers have also introduced various attention mechanisms, yielding breakthrough results. For instance, Wang et al. \cite{wang2020micro} proposed a micro-attention mechanism in conjunction with residual networks. This mechanism calculates attention maps using multi-scale features from residual networks, focusing on facial regions of interest and AUs to recognize micro-expressions. Yang et al. \cite{yang2021merta} proposed a convolutional neural network based on an attention mechanism that embeds static facial key points, dynamic information, and expression-related semantic attributes into the attention mechanism to obtain more differentiated visual representations. Wang et al. \cite{wang2023shallow} proposed a shallow multi-branch attention convolutional neural network that integrates attention mechanisms with multiple parallel feature processing branches to improve micro-expression recognition accuracy. Wang et al. \cite{wang2024htnet} introduced a hierarchical Transformer architecture that identifies key regions of facial muscles through a hierarchical structure to enhance micro-expression recognition accuracy.

However, these methods either rely on fixed feature patch sizes, lacking dynamic adaptability, or overlook the powerful representation capabilities of multi-scale feature synergy, resulting in insufficient detail feature capture and inadequate representation of complex micro-expression variations. Therefore, this paper proposes an Adaptive Hierarchical Multi-Scale Attention Mechanism (AHMSA-Net). Through a hierarchical framework with adaptive downsampling, AHMSA-Net dynamically adjusts the size of optical flow feature maps to capture subtle changes in micro-expressions at different granularities (fine and coarse). With a multi-scale attention mechanism, AHMSA-Net can integrate information from different scales (channel and spatial), thereby comprehensively improving the accuracy of micro-expression recognition.

\section{The Proposed Method}
\label{sec:method}

In this section, we first provide a brief overview of the overall framework of AHMSA -Net. We then give a detailed description of the two modules of AHMSA-Net: the adaptive hierarchical attention framework and the multi-scale attention module.

\subsection{Data Preprocessing}

For the four major micro-expression databases (SMIC, SAMM, CASMEII, CASM E\^{}3), we use the Dlib toolkit to locate facial landmarks and crop the images accordingly. Additionally, we extract the onset and apex frames from each database based on the annotation files and perform facial alignment on these frames using the facial landmarks.

\subsection{Three-Dimensional Optical Flow Feature Extraction}
\label{subsec:flow}

Optical flow methods are typically used to describe subtle muscle movements in the face \cite{li2020deep}, making them well-suited for capturing changes in micro-expression facial movements. Compared to other features, such as texture features, optical flow features can effectively reduce domain differences between different databases, which is crucial for improving cross-database micro-expression recognition performance \cite{khor2018enriched}. 

In this paper, we use the Total Variation-L1 optical flow (TV-L1) method to calculate the optical flow between the onset and apex frames. This method is suitable for describing small displacement changes between consecutive frames. We use $u$ and $v$ to denote the horizontal and vertical components of the optical flow field, respectively, to describe the motion information between the onset and apex frames. The authors of \cite{liong2019shallow} further extracted optical strain ($os$) from the optical flow, which helps in more accurately describing fine-scale movements. Therefore, we calculate the optical strain ($os$) as the third dimension of optical flow information based on $u$ and $v$. As shown in equation (1):
\begin{equation}
os=\dfrac{1}{2}\left[ \nabla O_{f}+\left( \nabla O_{f}\right) ^{T}\right]
\end{equation}

where $O_{f}=\left[ u,v\right] ^{T}$ is the optical flow vector, including horizontal and vertical components, and $\nabla$ is the derivative of $O_{f}$. And $u$ denotes the horizontal component of the optical flow field. $v$ denotes the vertical component of the optical flow field. $os$ denotes optical strain.

Subsequently, we detect the coordinates of five key facial regions from the apex frame of the micro-expression images: left eye, right eye, nose, left lip, and right lip. This step is performed using MediaPipe technology \cite{lugaresi2019mediapipe}. Using these coordinates, we map, crop, and recombine the three-dimensional optical flow maps to obtain the optical flow feature maps for these key facial regions. All optical flow feature maps are then unified to the size of $H_{flow} \times W_{flow} \times 3$ and used as input data for AHMSA-Net.

\subsection{Adaptive Hierarchical Attention Framework}
\label{subsec:adaptive}

\begin{figure}[h]
\centering
\includegraphics[width=\textwidth]{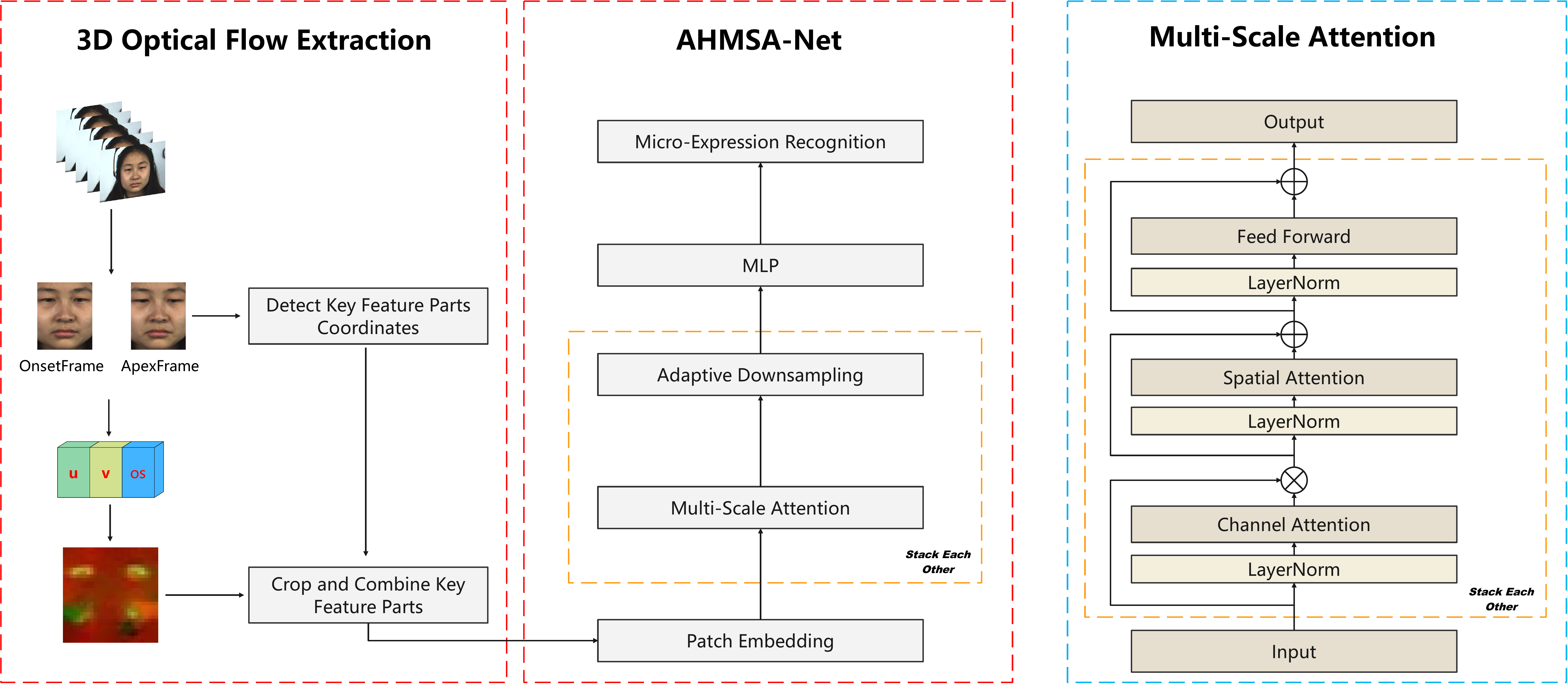}
\caption{The overall framework for micro-expression recognition. The micro-expression recognition framework highlighted in the red box is divided into two parts: the three-dimensional (3D) optical flow extraction part and the AHMSA-Net part. The multi-scale attention mechanism highlighted in the blue box consists of three stacked modules: the channel attention module, the spatial attention module, and the feed-forward module. Additionally, each module undergoes LayerNorm processing before data input.}\label{fig1}
\end{figure}

As shown in Figure~\ref{fig1}, the adaptive hierarchical attention framework of AHMSA-Net consists of two stacked modules: multi-scale attention and adaptive downsampling. The details of the multi-scale attention will be described in Section~\ref{subsec:multi}. This section focuses on the construction of the hierarchical attention framework with adaptive downsampling.

In AHMSA-Net, the optical flow feature map with dimensions $H_{flow} \times W_{flow} \times 3$ first undergoes Patch\_Embedding, converting the entire optical flow feature map into a sequence of optical flow feature patches with dimensions $P \times P \times 3$. The processed tensor $X \in \mathbb{R}^{B,C,H,W}$ is then used as input to the hierarchical framework, where $B$ denotes Batch Size, $C$ denotes Channels, $H$ is $H_{flow}/P$, and $W$ is $W_{flow}/P$. At lower levels, the number of optical flow feature patches is higher, while at higher levels, the number of patches is reduced. At the top level, there is only one optical flow feature patch, which is used for MLP prediction classification. The levels are connected through adaptive downsampling, as described in Equation (2):
\begin{equation}
X_{i+1} = AdaptiveMaxPool(LayerNorm(Conv(X_{i}))
\end{equation}

where $X_i$ is the patch tensor at the $i$-th layer, and $X_{i+1}$ is the processed result tensor at the $i+1$-th layer. Specifically, connections between layers first pass through a $3 \times 3$ convolutional layer to aggregate channel feature information, followed by layer normalization (LN) to improve model stability and convergence speed. Finally, adaptive max pooling dynamically adjusts the feature patch size after aggregation, enabling the network to retain important features while reducing redundant information. The layer normalization formula is given in Equation (3):
\begin{equation}
x_{i+1}=\dfrac{x_{i}-\mu }{\sqrt{\delta ^{2}+\varepsilon }}\cdot \gamma+\beta
\end{equation}

where $x_i$ is an element in the tensor $X$, $x_{i+1}$ is the result after processing in the LN layer, $\mu$ is the mean, $\delta ^{2}$ is the variance, $\varepsilon$ is a small constant for numerical stability, and $\gamma$ and $\beta$ are both learnable parameters for scaling and panning.

The hierarchical framework aims to help AHMSA-Net integrate features at different granularities. Specifically, lower-level attention mechanisms focus on fine-grained features of local regions, while higher-level attention mechanisms focus on coarse-grained features of global regions. This hierarchical attention framework effectively captures and integrates features at various granularities, improving micro-expression recognition.

\subsection{Multi-Scale Attention Mechanism}
\label{subsec:multi}

\begin{figure}[h]
\centering
\includegraphics[width=0.9\textwidth]{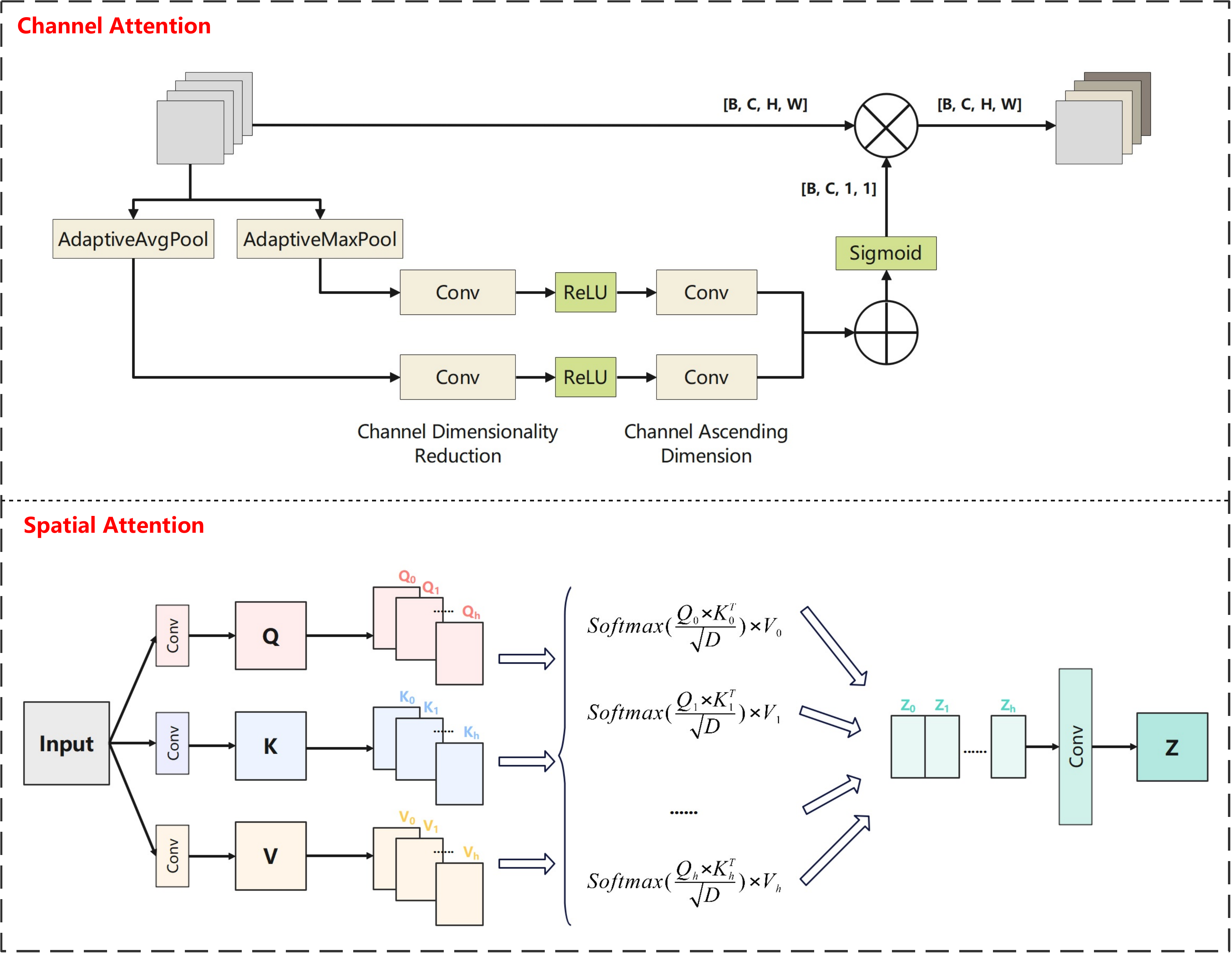}
\caption{Internal details of the channel attention module and spatial attention module.}\label{fig2}
\end{figure}

As shown in Figure~\ref{fig1}, the core of AHMSA-Net's multi-scale attention mechanism consists of three stacked components: the channel attention module, the spatial attention module, and the feed-forward module. The Feed Forward layer core is two $1 \times 1$ convolutional kernels, which help to enhance the model's ability of nonlinear transformation and linear combination of features. Additionally, each module undergoes LayerNorm before input to ensure model stability and convergence speed. This section will detail the core modules of the multi-scale attention mechanism: the channel attention module and the spatial attention module, as shown in Figure~\ref{fig2}.

\textbf{Channel Attention Module: }The channel attention module focuses on enhancing the important channels of each optical flow feature patch and suppressing noise and redundant information. As depicted in Figure~\ref{fig2}, the channel attention mechanism performs adaptive average pooling and adaptive max pooling on the input feature patches to obtain global average and global maximum information (representing significant features) for each channel. This information is then transformed through two $1 \times 1$ convolutional layers with ReLU activation functions. The outputs of these two pooling operations are summed and passed through a Sigmoid activation function to generate weights for each channel. These weights are then multiplied with the input feature patches on a per-channel basis, dynamically adjusting the importance of each channel to enhance the model's recognition performance.

\textbf{Spatial Attention Module: }The spatial attention module focuses on different optical flow feature patches at each layer to capture complex interaction information between features. Since micro-expression movements may occur simultaneously in different facial regions, studying the correlations between different optical flow feature patches helps improve the accuracy of micro-expression recognition. As shown in Figure~\ref{fig2}, the weighted optical flow feature patches, denoted as $Input \in R^{B, Heads, N, D}$ after channel attention weighting, are transformed into three components: Q, K, and V through $1 \times 1$ convolution. Subsequently, scaled dot-product attention is applied to Q, K, and V, as shown in Equation (4). The results $Z_i$, for $i \in [1, \ldots, h]$, are concatenated and processed through a $1 \times 1$ convolutional layer to produce the final result $Z$.
\begin{equation}
Z_i = softmax(\dfrac{Q_{i}K_{i}^{T}}{\sqrt{D}}) \dot V_i
\end{equation}

where, $B$ represents the batch size, $Heads$ denotes the number of attention heads, $N$ is the length of the patch sequence, $D$ is the input dimension, $Q$ represents the query matrix used to query relevant information, $K$ represents the key matrix paired with the query matrix to compute the match between queries, and $V$ is the value matrix containing the actual information, weighted according to the match.

Finally, the multi-scale attention and adaptive downsampling modules are stacked each other to form the core of AHMSA-Net. At the highest level of the AHMSA-Net hierarchical framework, adaptive downsampling reduces the size of the optical flow feature patches to $1 \times 1$ and extracts the complete feature map. The extracted feature map is then input into the MLP layer for the micro-expression recognition task.

\subsection{Loss Function}
\label{subsec:loss}

In this paper, we use the cross-entropy loss function, and the calculation formula of the cross-entropy loss $L$ is shown in Equation (5):

\begin{equation}
L =-\dfrac{1}{N}\sum ^{N}_{i=1}\sum ^{C}_{j=1}y_{ij}\log \left( \widehat{y}_{ij}\right)
\end{equation}

Where, $N$ is the number of samples, $C$ is the number of classes, $y_{ij}$ is the true label, indicating that the $i$-th element of the sample belongs to the $j$-th class, and $\hat{y}_{ij}$ is the predicted probability, indicating the probability that the $i$-th sample belongs to the $j$-th class.

\section{Experiments}
\label{sec:exp}

\subsection{Databases}
\label{subsec:data}

\begin{table}[h]
\renewcommand{\arraystretch}{1}
\centering
\resizebox{\textwidth}{!}{
\begin{tabular}{ccccc}
\hline
Database  & SMIC  & SAMM  & CASMEII  & CAS(ME)\^{}3 \\ \hline
Subject & 16  & 28  & 24  & 100\\ \hline
Samples & 164  & 133  & 145  & 943\\ \hline
Frame rate & 100  & 200  & 200  & 30\\ \hline
Optical flow resolution & 28 $\times$ 28  & 28 $\times$ 28  & 28 $\times$ 28  & 28 $\times$ 28 \\ \hline
Negative & 70  & 92  & 88  & 508 \\
Positive & 51  & 26  & 32  &64 \\
Surprise & 43   & 15  & 25  &201 \\ \hline
Onset index & $\checkmark$ & $\checkmark$  & $\checkmark$  & $\checkmark$ \\
Offset index & $\checkmark$  & $\checkmark$  & $\checkmark$  & $\checkmark$ \\
Apex index & $\times$  & $\checkmark$  & $\checkmark$  & $\checkmark$ \\  
\hline
\end{tabular}
}
\caption{Details of all databases. SMIC means SMIC-HS and CASME\^{}3 means CASME\^{}3-Part A. In addition, although the SMIC database does not have an Apex index, the images labeled with the Offset index exhibit effects similar to the Apex Frame, which captures the moments of most noticeable micro-expression changes. In actual training, we use the Offset index as the Apex index.}\label{tab1}
\end{table}

Experiments were conducted based on the four databases: SMIC \cite{li2013spontaneous,pfister2011recognising,li2017towards,tran2021micro}, SAMM \cite{davison2016samm,davison2018objective,yap2020samm}, CASMEII \cite{yan2014casme}, and CASME\^{}3 \cite{li2022cas}. During the experiments, we combined the SAMM, SMIC, and CASMEII databases into a composite micro-expression database (3DB). Additionally, we performed model validation on the larger micro-expression database CASME\^{}3. In these databases, emotional categories are classified into three types: positive, negative, and surprised. The "positive" category includes "happy," the "negative" category includes "sad," "disgust," "contempt," "fear," and "anger," and the "surprised" category includes only "surprised," as shown in Table~\ref{tab1}.

\subsection{Experimental Environment Information}
\label{subsec:envir}

\subsubsection{Hardware and Software Settings}
\label{subsubsec:hard}

In order to ensure the reliability of the experimental results, we conducted all the experiments in a unified environment, and the detailed experimental environment setting information is shown in Table~\ref{tab2}:

\begin{table}[h]
\renewcommand{\arraystretch}{1}
\centering
\begin{tabular}{l@{\hskip 3cm}l}
\hline
Item & Configuration\\
\hline
GPU & NVIDIA GeForce GTX 4090D\\
Operating System & Ubuntu18.04\\
Cuda & 11.3\\
Programming-Env & Python3.8\\
\hline
\end{tabular}
\caption{Hardware and software configuration for experiments.}\label{tab2}
\end{table}

\subsubsection{Hyperparameter Settings}
\label{subsubsec:hyper}

To provide a comprehensive understanding of the model's configuration, we detail the hyperparameters used in the experiments, as shown in Table~\ref{tab3}.

\begin{table}[h]
\renewcommand{\arraystretch}{1}
\centering
\begin{tabular}{l@{\hskip 1cm}l}
\hline
Parameter & Value\\
\hline
Epochs & 800 \\
Learning Rate & 0.000005 \\
Batch Size  & 256 \\
Optimizer & Adam \\
Number of Hierarchical Layers & 3 \\
Downsampling Scaling Factor & 2 \\
Number of Multi-Scale Attention Blocks Per Layer & (2,2,8) \\
$H_{flow} \times W_{flow}$ & 28 $\times$ 28 \\
$P \times P$ & 7 $\times$ 7 \\
$Heads$ & 3 \\
\hline
\end{tabular}
\caption{Hyperparameters settings.}\label{tab3}
\end{table}

\subsection{Evaluation Metrics}

In this paper, to ensure consistency and comparability, we adopted the Composite Database Evaluation (CDE) protocol \cite{see2019megc}. Additionally, Leave-One-Subject-Out (LOSO) cross-validation is employed to establish training-testing splits. This method entails holding out each subject group as the testing set while utilizing all other samples for training. 

Based on the sample counts of different classes reported in Table~\ref{tab1}, it is evident that these databases exhibit a phenomenon of imbalanced class sample distribution. To address this imbalance effectively \cite{le2015spontaneous}, performance is evaluated using two balanced metrics.

\subsubsection{Unweighted F1-score (UF1)}

The Unweighted F1-score ($UF1$), also known as the macro-average $F1$-score, is a metric that is commonly used to evaluate performance in multi-class classification tasks with imbalanced class distributions. To calculate the $UF1$, compute the $F1$-score for each emotion class $c$. In addition, we also need to compute the False Positives ($FP$), True Positives ($TP$), and False Negatives ($FN$) for each class $c$ in all folds of the leave-one-subject-out (LOSO) cross-validation. The $TP_c$ represents the number of samples correctly classified as class $c$ and they actually belong to class $c$. The $FP_c$ denotes the number of samples incorrectly classified as class $c$ while they actually belong to a different class. The $FN_c$ indicates the number of samples that belong to class $c$ but were incorrectly classified as not belonging to class $c$. The total number of emotion classes is denoted as $C$. The formula for calculating $UF1$ is given in Equation (6):
\begin{eqnarray}
F1_{c} &=& \dfrac{2 \times TP_{c}}{2 \times TP_{c} + FP_{c} + FN_{c}} \nonumber\\ 
UF1 &=& \dfrac{1}{C} \sum_{c=1}^{C} {F1_c}
\end{eqnarray}

\subsubsection{Unweighted Average Recall (UAR)}

Unweighted Average Recall ($UAR$): $UAR$ is a metric that is particularly useful when evaluating the effectiveness of a model in the presence of imbalanced class ratios. To calculate the $UAR$, we first compute the True Positives ($TP_c$) for each emotion class $c$. Additionally, we calculate the total number of samples ($n_c$) in each class. The total number of emotion classes is still denoted as $C$. The formula for calculating $UAR$ is given in Equation (7):
\begin{eqnarray}
ACC_{c} &=& \dfrac{TP_{c}}{n_{c}} \nonumber\\ 
UAR &=& \dfrac{1}{C} \sum_{c=1}^{C} ACC_{c} 
\end{eqnarray}

Both metrics offer a balanced assessment, ensuring that an approach's performance across all classes is equally evaluated. This reduces the risk of an approach being biased towards certain classes and provides a more comprehensive evaluation.

\subsection{Comparison with State-of-the Art Methods}
\label{subsec:compare}

To validate the effectiveness of the AHMSA-Net model, we conducted quantitative evaluation experiments on composite micro-expression databases (3DB), as well as the CAS(ME)\^{}3 micro-expression database. Additionally, we compared the results with several state-of-the-art methods. We also present the confusion matrices of our proposed method on each database and provide an in-depth analysis. The results and analysis will be detailed in the following sections.

\begin{table}[h]
\renewcommand{\arraystretch}{2}
\centering
\resizebox{\textwidth}{!}{
\begin{tabular}{cccccccccc}
\hline
\multirow{2}{*}{Year} & \multirow{2}{*}{Approachs} & \multicolumn{2}{c}{Full} & \multicolumn{2}{c}{SMIC} & \multicolumn{2}{c}{SAMM} & \multicolumn{2}{c}{CASMEII} \\ \cline{3-10} 
& & UF1  & UAR  & UF1  & UAR  & UF1  & UAR  & UF1  & UAR  \\ \hline
\textbf{Handcrafted} & & & & & & & & \\
2015 & LBP-TOP \cite{wang2015lbp} & 0.5882  & 0.5785  & 0.2000  & 0.5280  & 0.3954  & 0.4102  & 0.7026  & 0.7429  \\
2018 & Bi-WOOF \cite{liong2018less} & 0.6296  & 0.6227  & 0.5727  & 0.5829  & 0.5211  & 0.5139  & 0.7805  & 0.8026  \\ \hline

\textbf{Deep Learning} & & & & & & & & \\
2023 & Dual-ATME \cite{zhou2023dual} & 0.6790 & 0.6800 & 0.6460 & 0.6580 & 0.5620 & 0.5380 & 0.6460 & 0.6580 \\
2023 & SMBANet \cite{wang2023shallow} & 0.7440  & 0.7459  & 0.6914  & 0.6934  & 0.6021  & 0.6068  & 0.8970  & 0.8920  \\ 
2023 & BDCNN with CE loss \cite{chen2022block} & 0.7470  & 0.7225  & 0.6317  & 0.6233  & 0.7572  & 0.7186  & 0.8740  & 0.8491  \\
2023 & IncepTR \cite{zhou2023inceptr} & 0.7530  & 0.7460  & 0.6550  & 0.6500  & 0.6910  & 0.6940  & \textcolor{red}{0.9110}  & 0.8960  \\ 
2024 & CoTDPN \cite{yang2024micro} & 0.7424 & 0.7648 & 0.6321 & 0.6740 & 0.7367 & 0.7539 & 0.8015 & 0.7931 \\
2024 & LAENet \cite{gan2024laenet} & 0.7568  & 0.7405  & 0.6620  & 0.6523  & 0.6814  & 0.6620  & 0.9101  & \textcolor{red}{0.9119}  \\ \hline

2024 & AHMSA-Net (ours) & \textcolor{red}{0.8040}  & \textcolor{red}{0.7821}  & \textcolor{red}{0.7298}  & \textcolor{red}{0.7181} & \textcolor{red}{0.8803} & \textcolor{red}{0.8524}  & 0.8253  & 0.8029  \\ \hline
\end{tabular}
}
\caption{Performance Comparison of Various Methods on Composite Micro-expression Database (SAMM, SMIC, CASMEII). Red data indicates the best result in the current column.}\label{tab4}
\end{table}

\begin{table}[h]
\renewcommand{\arraystretch}{1}
\centering
\begin{tabular}{c@{\hskip 1.5cm}c@{\hskip 1.5cm}c@{\hskip 1.5cm}c}
\hline
\multirow{2}{*}{Year} & \multirow{2}{*}{Approachs} & \multicolumn{2}{c}{CAS(ME)\^{}3} \\ \cline{3-4} 
& & UF1  & UAR \\ \hline
2020 & RCN-A \cite{xia2020revealing} & 0.3928  & 0.3893  \\ 
2022 & FeatRef \cite{zhou2022feature} & 0.3493  & 0.3413  \\ 
2023 & $\mu$-BERT \cite{nguyen2023micron} & 0.5604  & 0.6125  \\ 
2024 & HTNet \cite{wang2024htnet} & 0.5767  & 0.5415  \\ \hline
2024 & AHMSA-Net (Ours) & \textcolor{red}{0.8120}  & \textcolor{red}{0.7708}  \\ \hline
\end{tabular}
\caption{Performance Comparison of Various Methods on CAS(ME)\^{}3 Micro-expression Database. Red data indicates the best result in the current column.}\label{tab5}
\end{table}

\subsubsection{Compared to Handcrafted Methods}
\label{subsubsec:hand}

As shown in Table~\ref{tab4}, we selected LBP-TOP and Bi-WOOF as representatives of handcrafted feature extraction methods on the 3DB database. These two methods classify micro-expressions based on appearance features and geometric features, respectively, and both methods use SVM as the classifier. In contrast, our proposed AHMSA-Net method shows a significant performance improvement in both UF1 and UAR on the composite database (3DB). Specifically, compared to the best-performing handcrafted method Bi-WOOF, AHMSA-Net improves the overall UF1 score by 17.44\% and the overall UAR score by 15.94\% on the 3DB database.

\subsubsection{Compared to Deep Learning Methods}
\label{subsubsec:deep}

As shown in Table~\ref{tab4} and Table~\ref{tab5}, our AHMSA-Net outperforms other deep learning methods by a wide margin.

From Table~\ref{tab4}, we selected some recent methods as comparison methods on the 3DB database, including Dual-ATME, SMBANet, BDCNN with CE loss, IncepTR, CoTDPN, and LAENet. In comparison, our proposed AHMSA-Net achieves UF1 and UAR of 0.8040 and 0.7821, respectively, on the complete composite database, representing an approximate 5\% improvement over the previous state-of-the-art methods. It is noteworthy that the LAENet method achieved a UAR of 91.19\% on the CASMEII database, but only 65.23\% and 66.20\% on the SMIC and SAMM databases, with a difference of 25.96\% and 24.99\%, respectively. This phenomenon is even more pronounced in the handcrafted methods LBP-TOP and Bi-WOOF. Some deep learning methods, such as IncepTR, also exhibit this issue. We consider that the high recognition accuracy on the CASMEII database by these methods may be due to overfitting during feature learning on the CASMEII database. As the database uses a LOSO cross-validation configuration, the models mainly learn features from the CASMEII database, leading to a decline in performance on other databases. In contrast, our proposed AHMSA-Net model achieves more balanced experimental results across all databases, demonstrating its stronger adaptability to various databases.

Moreover, we validated the performance of AHMSA-Net on the larger CASME\^{}3 micro-expression database. The CASME\^{}3 database, compared to SMIC, SAMM, and CASMEII, has more samples and a more scientifically induced micro-expression elicitation method. Therefore, the experimental results on this database better reflect the true performance of the model, as detailed in Table~\ref{tab5}. Specifically, we selected RCN-A, FeatRef, $\mu$-BERT, and HTNet as comparison methods on the CASME\^{}3 database. In comparison, our proposed AHMSA-Net achieves UF1 and UAR of 0.8120 and 0.7708, respectively, on the CASME\^{}3 database, representing an approximate 24\% and 16\% improvement over the previous state-of-the-art methods.

In summary, these results highlight the effectiveness of the AHMSA-Net design and its significant advantages in micro-expression recognition tasks.

\subsubsection{Confusion Matrix Analysis}
\label{subsubsec:cm}

\begin{figure}[h]
\centering
\includegraphics[width=\textwidth]{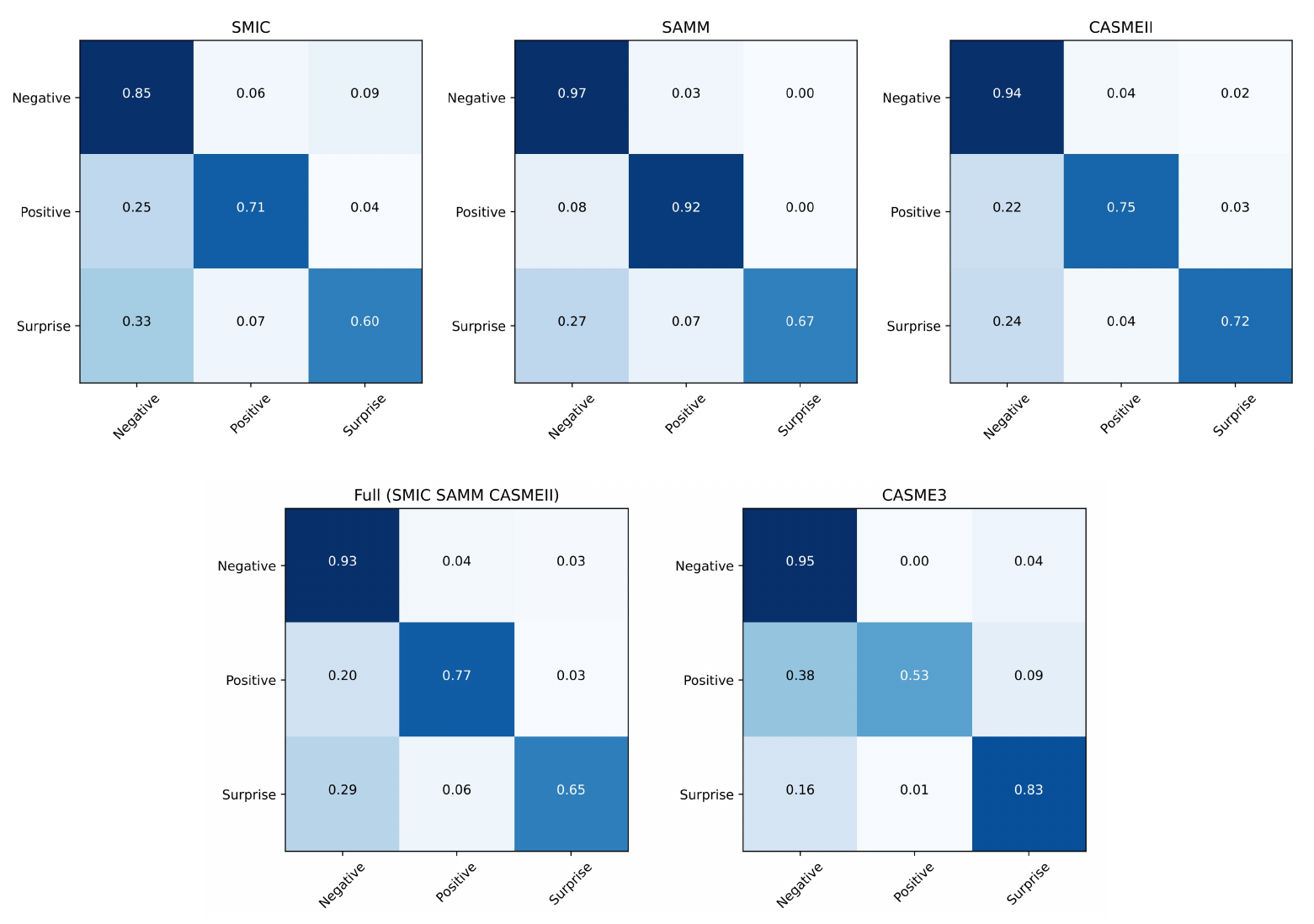}
\caption{Confusion Matrix for All Micro-expression Databases.}\label{fig3}
\end{figure}

Figure~\ref{fig3} presents the confusion matrices of AHMSA-Net on all micro-expression databases, showing the accuracy achieved for each emotion category. In the composite micro-expression database (3DB), AHMSA-Net achieves accuracy rates of 0.93, 0.77, and 0.65 for the Negative, Positive, and Surprise categories, respectively; in the larger CASME\^{}3 micro-expression database, AHMSA-Net achieves accuracy rates of 0.95, 0.53, and 0.83 for the Negative, Positive, and Surprise categories, respectively. The high accuracy rate for the Negative category can be attributed to the greater number of training samples in these databases, but the lower number of training samples for the other two categories often results in lower accuracy.

It is worth noting that the SMIC database, due to its use of a lower frame rate to capture micro-expressions, introduces background noise such as flickering lights and shadows, which leads to the lowest accuracy rates for the three categories in the composite micro-expression database (3DB), with accuracy rates of 0.85, 0.71, and 0.60 for the Negative, Positive, and Surprise categories, respectively. Additionally, the accuracy rate for the Positive category in the CASME\^{}3 database is also unsatisfactory, only 0.53, with 38\% of the samples incorrectly classified into the Negative category. We consider this to be the result of multiple factors, including overlapping category features and sample imbalance.

\subsection{Ablation Study}
\label{subsec:abl}

This section provides an in-depth analysis of how key parameters affect the AHMSA-Net model. Specifically, we examine the impact of Batch Size, the number of hierarchical layers, and the number of Multi-Scale attention blocks per layer. For each ablation experiment, we use the UF1 and UAR metrics for evaluation.

\subsubsection{Impact of Batch Size}
\label{subsubsec:batch}

\begin{figure}[h]
\centering
\includegraphics[width=\textwidth]{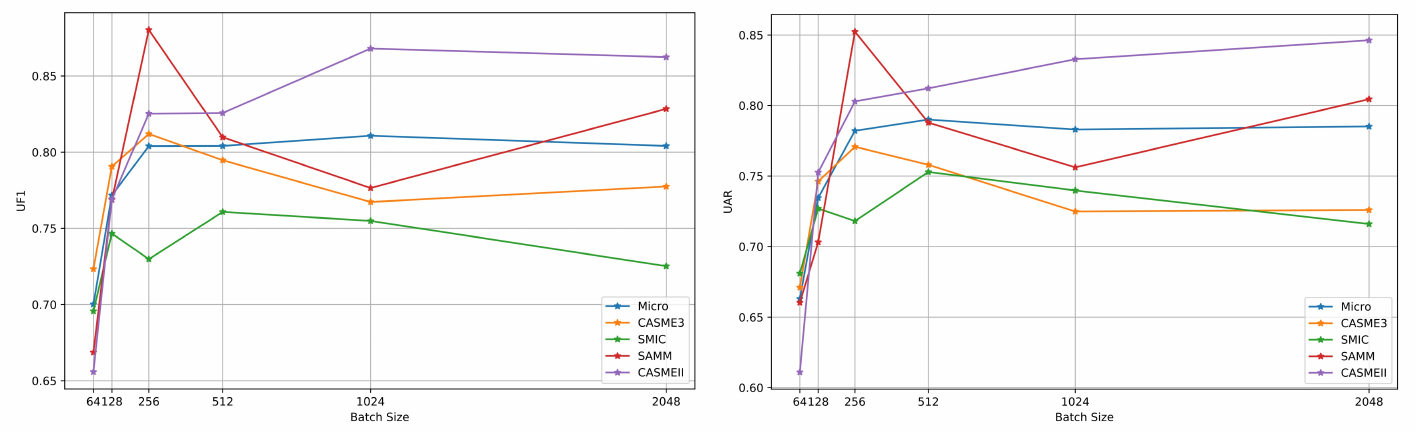}
\caption{Impact of Batch Size.}\label{fig4}
\end{figure}

We first investigate the effect of Batch Size on overall accuracy. The experimental results of AHMSA-Net on the Composite Data Set (3DB) and CASME3 Data Set (UF1 and UAR values) are shown in Figure~\ref{fig4}. Batch Size is a critical parameter in AHMSA-Net training, and an appropriate Batch Size can maximize the model’s performance.

From Figure~\ref{fig4}, it is evident that the performance of the AHMSA-Net model is optimal when the Batch Size is 256. When the Batch Size is smaller than 256, the performance across all databases is suboptimal and does not meet expectations. Conversely, when the Batch Size exceeds 256, there is a noticeable inflection point in recognition accuracy for the 3DB Composite Database and CASME3, accompanied by a significant decline. Although there is a slight improvement in the SMIC database, overall, a Batch Size of 256 remains the best choice. Notably, CASMEII is the only database that continues to improve with larger Batch Sizes. This phenomenon is similar to the overfitting observed in other methods discussed in section~\ref{subsubsec:deep}, further emphasizing the importance of selecting an appropriate Batch Size to balance model adaptability.

\subsubsection{Impact of Number of Hierarchical Layers \& Downsampling Scale Factor}
\label{subsubsec:layer}

We further analyze the impact of the number of hierarchical layers and downsampling scale factor on AHMSA-Net model performance.

First, the number of layers is related to the dimensions of the optical flow feature map $H_{flow} \times W_{flow}$, the size of the feature blocks $P \times P$, and the downsampling scale factor. To avoid introducing excessive uncertainty in parameter quantity, each layer's patch blocks in the hierarchical framework are set to square shapes rather than irregular shapes. By adjusting the $Ratio = H_{flow}/P$ (or $W_{flow}/P)$ in relation to the downsampling scale factor, we explored the suitable number of layers. In the initial setting, the $Ratio$ is $28/7 = 4$, and the downsampling scale factor is 2, allowing for three hierarchical layers through two scaling operations.

Further, when we fixed the number of layers to 3 and experimented with different downsampling scale factors (such as 3, 4, 5, etc.), we found that the network's Batch Size was limited to 32 and could not reach 256. This limitation is likely due to the fact that with a downsampling factor of 3, $H_{flow} \times W_{flow}$ should be $36 \times 36$, increasing the number of patches in the lowest layer to 81, which significantly adds to the network's complexity. Notably, 36 is the least common multiple of $2^2$ and $3^2$. If more downsampling factors are used, this value would be even larger. Therefore, considering the impact of Batch Size on model performance as discussed in section~\ref{subsubsec:batch}, we fixed the downsampling scale factor at 2.

\subsubsection{Impact of Number of Multi-Scale Attention Blocks Per Layer}
\label{subsubsec:MSA}

\begin{figure}[h]
\centering
\includegraphics[width=\textwidth]{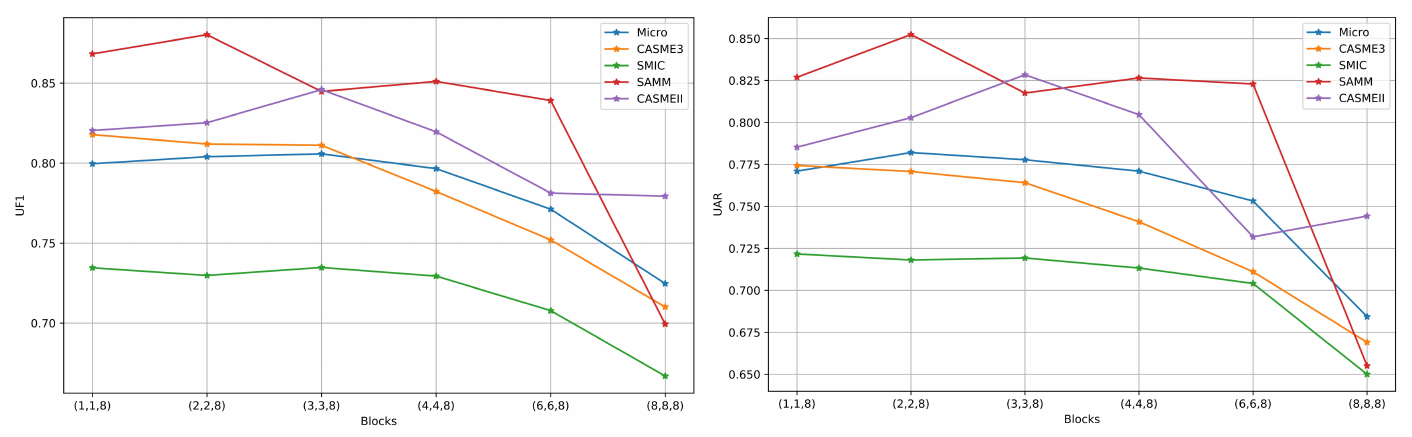}
\caption{Impact of Number of Multi-Scale Attention Blocks.}\label{fig5}
\end{figure}

In the hierarchical framework of AHMSA-Net, the core component of each layer is the multi-scale attention. In this section, we explore the impact of the number of multi-scale attention blocks per layer on the model's performance. Specifically, the number of blocks in each layer of the initial network is set to (2, 2, 8) from low to high. We drew inspiration from the parameter setting in \cite{wang2024htnet}, where the number of Transformer blocks in the highest layer is set to 8. Therefore, we only investigate the effect of the number of multi-scale attention blocks in the lower layers on the performance of the AHMSA-Net network, considering the configurations (1, 1, 8), (3, 3, 8), (4, 4, 8), (6, 6, 8) and (8, 8, 8). The experimental results are shown in Figure~\ref{fig5}.

From Figure~\ref{fig5}, it can be seen that using a configuration with fewer multi-scale attention blocks, specifically (2, 2, 8), allows AHMSA-Net to achieve optimal performance on the Micro-Expression Conformity Database (3DB) and CASME\^{}3 databases.

\section{Conclusion}
\label{sec:con}

In this paper, we propose an Adaptive Hierarchical Multi-Scale Attention Network for Micro-Expression Recognition (AHMSA-Net). Specifically, we first utilize the onset and apex frames of micro-expression sequences to extract three-dimensional optical flow maps, including horizontal flow, vertical flow, and flow strain. Subsequently, these optical flow feature maps are input into the AHMSA-Net, which consists of two main components: an adaptive hierarchical framework and a multi-scale attention mechanism. Based on the adaptive downsampling hierarchical attention framework, AHMSA-Net can dynamically adjust the size of optical flow feature maps at each layer, thereby capturing subtle changes in micro-expressions at different granularities (fine and coarse). Leveraging the multi-scale attention mechanism, AHMSA-Net can learn micro-expression motion information by fusing information from different scales (channel and spatial). These two modules work in tandem to comprehensively improve the accuracy of micro-expression recognition. Experimental results demonstrate the superiority of our proposed method.

In future work, we will further optimize the model's architecture, aiming to transform some of the fixed parameters discussed in the ablation experiments into dynamic parameters. Additionally, we plan to explore more methods in database preprocessing, anticipating further improvements in the quality of micro-expression databases. Lastly, we hope to apply our model to a wider range of visual tasks.

\bibliographystyle{elsarticle-num}
\bibliography{refs}

\end{document}